\documentclass[runningheads,a4paper]{llncs}

\usepackage{amssymb}
\usepackage{graphicx}
\usepackage{amsmath}
\usepackage{caption} 
\usepackage{url}
\urldef{\mailsa}\path|{srikanth.cherla.1, son.tran.1,|
\urldef{\mailsb}\path|t.e.weyde, a.garcez}@city.ac.uk|    
\newcommand{\keywords}[1]{\par\addvspace\baselineskip
\noindent\keywordname\enspace\ignorespaces#1}

\usepackage{pgfplots}

\usepackage{tikz}
\usetikzlibrary{decorations.pathmorphing} % noisy shapes
\usetikzlibrary{fit}                    % fitting shapes to coordinates
\usetikzlibrary{backgrounds}    % drawing the background after the foreground
\usetikzlibrary{shapes, arrows}

\tikzstyle{matrx}=[rectangle, thick, minimum size=1.0cm, draw=gray!80,
                   fill=gray!20] 

\tikzstyle{background}=[rectangle, draw=gray!40, fill=gray!10, inner
                        sep=0.1cm, rounded corners=1mm]
\usepackage{subcaption} % For subfigures
\begin{document}

\mainmatter  % start of an individual contribution

\title{Generalising the Discriminative Restricted Boltzmann Machine}

\titlerunning{Generalising the Discriminative RBM}

\author{Srikanth Cherla \and Son N. Tran \and Tillman Weyde \and \\
    Artur d'Avila Garcez}
 
\authorrunning{Cherla et al.}

\institute{School of Mathematics, Computer Science and Engineering\\
    City University London \\
    Northampton Square, London EC1V 0HB. United Kingdom\\
\mailsa \mailsb\\}

\toctitle{Generalising the Discriminative Restricted Boltzmann Machine}
\tocauthor{Srikanth Cherla, Son N. Tran, Tillman Weyde, Artur d'Avila Garcez}
\maketitle

\begin{abstract}
We present a novel theoretical result that generalises the Discriminative
Restricted Boltzmann Machine (DRBM).
While originally the DRBM was defined assuming the $\{0, 1\}$-Bernoulli
distribution in each of its hidden units, this result makes it possible to 
derive cost functions for variants of the DRBM that utilise other
distributions, including some that are often encountered in the literature.
This is illustrated with the Binomial and $\{-1, +1\}$-Bernoulli distributions
here.
We evaluate these two DRBM variants and compare them with the original
one on three benchmark datasets, namely the MNIST and USPS digit
classification datasets, and the 20 Newsgroups document classification dataset.
%Results show that these two variants perform comparably to the original DRBM
%and in some cases better than it, thus presenting viable alternatives.
Results show that each of the three compared models outperforms the remaining
two in one of the three datasets, thus indicating that the proposed theoretical
generalisation of the DRBM may be valuable in practice.
\end{abstract}

    \section{Introduction}
    \label{sec:introduction}
    The Restricted Boltzmann Machine (RBM) is a generative latent-variable
    model which models the joint distribution of a set of input variables.
    It has gained popularity over the past decade in many applications,
    including feature learning \cite{Lee2009}, collaborative filtering 
    \cite{Salakhutdinov2007}, high-dimensional sequence modelling
    \cite{Sutskever2007}, and pretraining deep neural network 
    classifiers \cite{Mohamed2012}.
    One of its applications is as a standalone classifier, referred to as the
    Discriminative Restricted Boltzmann Machine (DRBM).
    As the name might suggest, the DRBM is a classifier obtained by carrying
    out discriminative learning in the RBM and it directly models the
    conditional distribution one is interested in for prediction.
    This bypasses one of the key problems faced in learning the parameters of
    the RBM generatively, which is the computation of the intractable
    \textit{partition function} (discussed in Section \ref{sec:rbm}).
    In the DRBM this partition function is cancelled out in the expression for
    the conditional distribution thus simplifying the learning process.
 
    In the original work that introduced the DRBM its cost function was
    derived under the assumption that each of its hidden units models the
    $\{0, 1\}$-Bernoulli distribution.
    We observe that while effort has gone into enhancing the performance of a
    few other connectionist models by changing the nature of their hidden
    units, this has not been attempted with the DRBM.
    So in this paper, we first describe a novel theoretical result that makes
    it possible to generalise the model's cost function.
    The result is then used to derive two new cost functions corresponding to
    DRBMs containing hidden units with the Binomial and $\{-1, +1\}$-Bernoulli 
    distributions respectively.
    These two variants are evaluated and compared with the original DRBM on the
    benchmark MNIST and USPS digit classification datasets, and the 20
    Newsgroups document classification dataset.
    We find that each of the three compared models outperforms the remaining
    two in one of the three datasets, thus indicating that the proposed
    theoretical generalisation of the DRBM may be valuable in practice.

    The next section describes the motivation for this paper and covers the
    related work that led to the ideas presented here.
    This is followed by an introduction to the Restricted Boltzmann Machine
    (RBM) in Section \ref{sec:rbm} and its discriminative counterpart, the
    Discriminative RBM (DRBM) in Section \ref{sec:drbm}. 
    The latter of these is the focus of this paper.
    Section \ref{sec:drbm_extensions} first presents the novel theoretical
    result that generalises this model, and then its two aforementioned
    variants that follow as a consequence.
    These are evaluated on three benchmark datasets, and the results discussed
    in Section \ref{sec:experiments}.
    Section \ref{sec:conclusions} presents a summary, together with potential
    extensions of this work.

    \section{Related Work}
    \label{sec:related_work}
    The subject of activation functions in has received attention from time to
    time in the literature.  
    Previous work, in general, has either introduced a new type of activation
    function \cite{Teh2001}\cite{Welling2004}\cite{Hinton2009}\cite{Nair2010},
    carried out a comparative evaluation \cite{Gupta1992}\cite{Karlik2011}, or 
    re-evaluated the significance of an existing activation function in a new 
    context \cite{Glorot2010}\cite{Glorot2011}.
    In the case of feedforward neural networks, while \cite{Gupta1992}
    concluded in favour of the Logistic Sigmoid activation function, recent
    work in \cite{Glorot2010} found a drawback in its use with deep
    feedforward networks and suggested the Hyperbolic Tangent 
    function as a more suitable alternative to it.
    Even more recently, \cite{Glorot2011} highlighted the biological
    plausibility of the Rectified Linear activation function and its potential
    to outperform both the aforementioned activations without even the need to
    carry out unsupervised pre-training.
    
    The RBM was originally defined as a model for binary-valued variables whose
    values are governed by the $\{0, 1\}$-Bernoulli distribution.
    An early variant of this model \cite{Freund1992} known as the
    \textit{Influence Combination Machine}, while still modelling binary data,
    employed units of values $\{-1, +1\}$. 
    In \cite{Welling2004}, an extension of the standard RBM to model
    real-valued variables in its visible layers was proposed with the aid of
    the Gaussian activation function.
    With the same goal of modelling non-binary variables, the rate-coded RBM
    \cite{Teh2001} was introduced in which both the visible and hidden layers
    could model integer-values by grouping the activations of sets of binary
    units all of which share the same values of weights. 
    This idea was extended in \cite{Nair2010} which introduced Rectified Linear
    activations for the hidden layer of the RBM. 
    In the context of topic modelling, the replicated softmax RBM
    \cite{Hinton2009} was introduced which models the categorical distribution
    in its visible layer.
    All these variants of the RBM have been shown to be better suited for
    certain specific learning tasks.

    It is often the case that a new type of activation function results in an
    improvement in the performance of an existing model or in a new insight
    into the behaviour of the model itself. 
    In the least, it offers researchers with the choice of a new modelling
    alternative.
    This is the motivation for the present work.
    As outlined above, effort has gone into enhancing the performance of a few 
    models in this way.
    We observe that it has not, however, been attempted with the DRBM.
    So here we address this by first introducing two new variants of the DRBM
    based on a novel theoretical result, and then comparing the performance of
    these and the original $\{0,1\}$-Bernoulli DRBM on benchmark datasets.

    \section{Restricted Boltzmann Machine}
    \label{sec:rbm}
\begin{figure}[htbp]
        \centering
        \begin{subfigure}{.48\textwidth}
        \centering
        % Definition for foreground rectangles. They have a smaller minimal
        % size for aesthetic reasons.
        \tikzstyle{matrx}=[rectangle, thick, minimum size=0.75cm, draw=gray!80,
                           fill=gray!20] 

        % Definition for background rectangles. Everything is drawn on
        % underlying gray rectangles with rounded corners.
        \tikzstyle{background}=[rectangle, draw=gray!40, fill=gray!10, inner
                                sep=0.1cm, rounded corners=1mm]

        \begin{tikzpicture}[>=latex,text height=1.5ex,text depth=0.25ex]
            % The various elements are conveniently placed using a matrix:
            \matrix[row sep=1.5cm,column sep=0.5cm] {

            % Hidden layer
            \node (h) [matrx, draw=none, fill=gray!10] 
                       {$\ \ \ \ \ \mathbf{h}\ \ \ \ \ $}; \pgfmatrixnextcell
            \\

            % Input layer
            \node (v) [matrx] 
                      {$\ \ \ \ \ \ \ \ \ \ \mathbf{v}\ \ \ \ \ \ \ \ \ \ $}; \pgfmatrixnextcell
            \\
            };
 
            % Hidden layer grouping
            \begin{pgfonlayer}{background}
                \node (h) [background, fit=(h)] {};
            \end{pgfonlayer}

            % Connections
            \draw [-, thick] (v) -- (h) 
                node[pos=0.5, right]{$W$};
            ;
        \end{tikzpicture}
        \caption{}
    \end{subfigure}
    \begin{subfigure}{.48\textwidth}
    \centering
        % Definition for foreground rectangles. They have a smaller minimal
        % size for aesthetic reasons.
        \tikzstyle{matrx}=[rectangle, thick, minimum size=0.75cm, draw=gray!80,
                           fill=gray!20] 

        % Definition for background rectangles. Everything is drawn on
        % underlying gray rectangles with rounded corners.
        \tikzstyle{background}=[rectangle, draw=gray!40, fill=gray!10, inner
                                sep=0.1cm, rounded corners=1mm]

        \begin{tikzpicture}[>=latex,text height=1.5ex,text depth=0.25ex]
            % The various elements are conveniently placed using a matrix:
            \matrix[row sep=1.5cm,column sep=0.5cm] {

            % Hidden layer
            \node (h) [matrx, draw=none, fill=gray!10] 
                       {$\ \ \ \ \ \mathbf{h}\ \ \ \ \ $}; \pgfmatrixnextcell
            \\

            % Visible layer
            \node (x) [matrx] 
                {$\ \ \ \ \ \ \ \ \mathbf{x}\ \ \ \ \ \ \ \  $}; \pgfmatrixnextcell
                \node (y) [matrx] {$\ \ \ \mathbf{y}\ \ \ $}; \pgfmatrixnextcell

            % Output layer
            \\
            };
 
            % Hidden layer grouping
            \begin{pgfonlayer}{background}
                \node (h) [background, fit=(h)] {};
            \end{pgfonlayer}

            % Input layer grouping
            \begin{pgfonlayer}{background}
                \node (v) [background, fit=(x) (y)] {};%,
            \end{pgfonlayer}

            % Connections
            \draw [-, thick] (x) -- (h) 
                node[pos=0.5, right]{$R$};
            \draw [-, thick] (y) -- (h) 
                node[pos=0.5, right]{$U$};
            ;
        \end{tikzpicture}
        \caption{}
    \end{subfigure}
        \caption{The architecture of an RBM (left) which models the joint
            distribution $p(\mathbf{v})$, and a DRBM (right) which models the
            conditional distribution $p(\mathbf{y}|\mathbf{x})$.} 
    \label{fig:rbm_drbm}
    \end{figure}
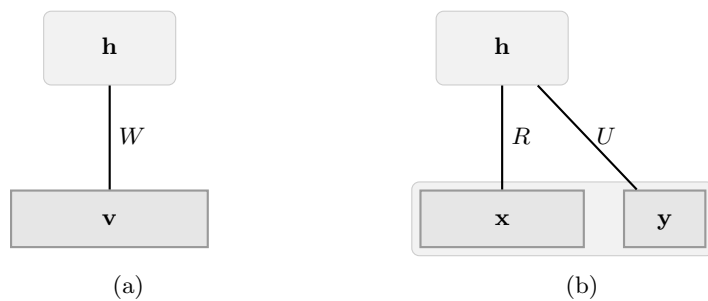

    The Restricted Boltzmann Machine (RBM) \cite{Smolensky1986} is an
    undirected bipartite graphical model. 
    It contains a set of visible units $\mathbf{v} \in \mathbb{R}^{n_v}$ and a
    set of hidden units $\mathbf{h} \in \mathbb{R}^{n_h}$ which make up its
    visible and hidden layers respectively.
    The two layers are fully inter-connected but there exist no connections
    between any two hidden units, or any two visible units. 
    Additionally, the units of each layer are connected to a bias unit whose
    value is always 1.
    The edge between the $i^{th}$ visible unit and the $j^{th}$ hidden unit is
    associated with a weight $w_{ij}$. 
    All these weights are together represented as a weight matrix $W \in
    \mathbb{R}^{n_v \times n_h}$.
    The weights of connections between visible units and the bias unit are
    contained in a visible bias vector $\mathbf{b} \in \mathbb{R}^{n_v}$.
    Likewise, for the hidden units there is a hidden bias vector $\mathbf{c}
    \in \mathbb{R}^{n_h}$. 
    The RBM is fully characterized by the parameters $W$, $\mathbf{b}$
    and $\mathbf{c}$.
    Its bipartite structure %(if one ignores the bias unit) 
    is illustrated in Figure \ref{fig:rbm_drbm}.
    
    The RBM is a special case of the Boltzmann Machine --- an energy-based model
    \cite{Lecun2006} --- which gives the joint probability of every possible pair
    of visible and hidden vectors via an energy function $E$, according to the
    equation
    \begin{equation}
        P(\mathbf{v}, \mathbf{h}) = \frac{1}{Z} 
            \mathrm{e}^{-E(\mathbf{v}, \mathbf{h})}
    \end{equation}
    where the ``partition function'', $Z$, is given by summing over all
    possible pairs of visible and hidden vectors
    \begin{equation}
        Z = \sum_{\mathbf{v}, \mathbf{h}} 
            \mathrm{e}^{-E(\mathbf{v}, \mathbf{h})}\ .
    \end{equation}
    and ensures that $P(\mathbf{v}, \mathbf{h})$ is a probability.
    The joint probability assigned by the model to the elements of a visible
    vector $\mathbf{v}$, is given by summing (marginalising) over all possible
    hidden vectors:
    \begin{equation}
        P(\mathbf{v}) = \frac{1}{Z} \sum_{\mathbf{h}}
            \mathrm{e}^{-E(\mathbf{v}, \mathbf{h})}
    \end{equation}
    In the case of the RBM, the energy function $E$ is given by
    \begin{equation}
        E(\mathbf{v}, \mathbf{h}) = -\mathbf{b}^{\top}\mathbf{v} -
            \mathbf{c}^{\top}\mathbf{h} -
            \mathbf{v}^{\top}\mathbf{W}\mathbf{h}\ .
    \end{equation}
    In its original form, the RBM models the Bernoulli distribution in its
    visible and hidden layers.
    The activation probabilities of the units in the hidden layer given the
    visible layer (and vice versa) are $P(\mathbf{h} = \mathbf{1} | \mathbf{v})
    = \sigma(\mathbf{c} + W^{\top} \mathbf{v})$ and $P(\mathbf{v} = \mathbf{1}
    | \mathbf{h}) = \sigma(\mathbf{b} + W \mathbf{h})$ respectively, where
    $\sigma(\mathbf{x})$ is the logistic sigmoid function $\sigma(\mathbf{x}) =
    (1+e^{-\mathbf{x}})^{-1}$ applied element-wise to the vector $\mathbf{x}$.

    Learning in energy-based models can be carried \textit{generatively}, by
    determining the weights and biases that minimize the overall energy of the
    system with respect to the training data.
    This amounts to maximizing the log-likelihood function $\mathcal{L}$ over
    the training data $\mathcal{V}$ (containing $N$ examples), which is given
    by 
    \begin{equation}
        \mathcal{L} = \frac{1}{N} \sum_{n=1}^{N} \log P(\mathbf{v}_n)
    \end{equation}
    where $P(\mathbf{v})$ is the joint probability distribution given by
    \begin{equation}
        P(\mathbf{v}) = \frac{e^{-E_{free}(\mathbf{v})}}{Z} ,
    \end{equation}
    with $Z = \sum_{\mathbf{v}}e^{-E_{free}(\mathbf{v})}$, and
    \begin{equation}
        E_{free}(\mathbf{v}) = -\log \sum_{\mathbf{h}}e^{-E(\mathbf{v},
            \mathbf{h})}\ . 
    \end{equation}
    The probability that the RBM assigns to a vector $\mathbf{v}_n$ belonging
    to the training data can be raised by adjusting the weights and biases to
    lower the energy associated with that vector and to raise the energy
    associated with others not in the training data.
    Learning can be carried out using gradient-based optimisation, for which
    the gradient of the log-likelihood function with respect to the RBM's
    parameters $\mathbf{\theta}$ needs to be calculated first.
    This is given by
    \begin{equation}
        \frac{\partial \mathcal{L}}{\partial \mathbf{\theta}} = 
        -\Biggl\langle \frac{\partial E_{free}}{\partial \mathbf{\theta}}
            \Biggr\rangle_0
        +\Biggl\langle \frac{\partial E_{free}}{\partial \mathbf{\theta}}
            \Biggr\rangle_\infty
    \end{equation}
    where $\langle \cdot \rangle_0$ denotes the average with respect to the
    data distribution, and $\langle \cdot \rangle_\infty$ that with respect to
    the model distribution.
    The former is readily computed using the training data $\mathcal{V}$, but
    the latter involves the normalisation constant $Z$, which very often cannot
    be computed efficiently as it involves a sum over an exponential number of
    terms.
    To avoid the difficulty in computing the above gradient, an efficiently
    computable and widely adopted approximation of the gradient was proposed in
    the Contrastive Divergence method \cite{Tieleman2008}.

    \section{The Discriminative Restricted Boltzmann Machine}
    \label{sec:drbm}
    The generative RBM described above models the joint probability
    $P(\mathbf{v})$ of the set of binary variables $\mathbf{v}$.
    For prediction, one is interested in a conditional distribution of the form
    $P(y|\mathbf{x})$.
    It has been demonstrated in \cite{Larochelle2008} how discriminative
    learning can be carried out in the RBM, thus making it feasible to use it
    as a standalone classifier.
    This is done by assuming one subset of its visible units to be inputs
    $\textbf{x}$, and the remaining a set of categorical units $\mathbf{y}$
    representing the class-conditional probabilities $P(y|\mathbf{x})$.
    This is illustrated in Figure \ref{fig:rbm_drbm}. 
    The weight matrix $W$ can be interpreted as two matrices $R \in
    \mathbb{R}^{n_i \times n_h}$ and $U \in \mathbb{R}^{n_c \times n_h}$, where
    $n_i$ is the input dimensionality and $n_c$ is the number of classes and
    $n_v = n_i + n_c$.
    Likewise, the visible bias vector $\mathbf{b} \in \mathbb{R}^{n_v}$ is also
    split into a set of two bias vectors --- a vector $\mathbf{a} \in
    \mathbb{R}^{n_i}$ and a second vector $\mathbf{d} \in \mathbb{R}^{n_c}$, as
    shown in Figure \ref{fig:rbm_drbm}.

    The posterior probability in this \textit{Discriminative} RBM can be
    inferred as
    \begin{equation} 
        P(\mathbf{y} |\mathbf{x})
        = \frac{\exp{(-E_{free}(\mathbf{x}, \mathbf{y}))}}
        {\sum_{\mathbf{y}^*}\exp{(-E_{free}(\mathbf{x}, \mathbf{y}^*))}}
    \end{equation}
    where $\mathbf{x}$ is the input vector, and $\mathbf{y}$ is the one-hot
    encoding of the class-label.
    The denominator sums over all class-labels $\mathbf{y}^*$ to make
    $P(\mathbf{y} |\mathbf{x})$ a probability distribution.
    In the original RBM, $\mathbf{x}$ and $\mathbf{y}$ together make up the
    visible layer $\mathbf{v}$.
    The model is learned discriminatively by maximizing the log-likelihood
    function based on the following expression of the conditional distribution:
    \begin{equation}
        \begin{aligned}
    P(y | \mathbf{x}) =
        &\frac{\exp(d_{y})
            \prod_j(1+\exp(\mathbf{r}_j^\top \mathbf{x} +
            u_{yj}+c_j))}
        {\sum_{y^{*}}\exp(d_{y^{*}})
            \prod_j(1+\exp(\mathbf{r}_j^\top \mathbf{x} +
            u_{y^{*}j}+c_j))} \ .
        \label{eq:drbm_cost}
    \end{aligned}
        \end{equation} 
    The gradient of this function, for a single input-label pair
    $(\mathbf{x}_i, y_i)$ with respect to its parameters $\mathbf{\theta}$ can
    be computed analytically.
    It is given by 
    \begin{equation}
        \begin{aligned}
        \frac{\partial\log P\left(y_i|\mathbf{x}_i\right)}
            {\partial\mathbf{\theta}}
        &= \sum_j \sigma\left(o_{yj}\left(\mathbf{x}_i\right)\right) 
            \frac{\partial o_{yj}\left(\mathbf{x}_i\right)}
            {\partial \mathbf{\theta}} \\
            &- \sum_{j, y^*}
                \sigma\left(o_{y^*j}\left(\mathbf{x}_i\right)\right)
                p\left(y^*|\mathbf{x}_i\right) 
                \frac{\partial o_{y^*j}\left(\mathbf{x}_i\right)}
                {\partial \mathbf{\theta}}
    \end{aligned}
        \end{equation}
    where $o_{yj}(\mathbf{x}) = c_j + \mathbf{r}^\top_{j}\mathbf{x} +
    u_{yj}$.
    Note that in order to compute the conditional distribution in
    (\ref{eq:drbm_cost}) the model does not have to be learned
    discriminatively, and one can also use the above generatively learned RBM
    as it learns the joint distribution $p(\mathbf{y}, \mathbf{x})$ from which
    $p(\mathbf{y}|\mathbf{x})$ can be inferred.

    \section{Generalising the DRBM}
    \label{sec:drbm_extensions}
    This section describes a novel generalisation of the cost function of the
    DRBM \cite{Larochelle2008}. 
    This facilitates the formulation of similar cost functions for variants of
    the model with other distributions in their hidden units that are commonly
    encountered in the literature.
    This is illustrated here first with the $\{-1, +1\}$-Bernoulli
    distribution and then the Binomial distribution.

        \subsection{Generalising the Conditional Probability}
        \label{subsec:drbm_generalisation}
        We begin with the expression for the conditional distribution
        $P(y|\mathbf{x})$, as derived in \cite{Larochelle2008}. 
        This is given by
        \begin{equation}
        \begin{aligned}
            P(y|\mathbf{x}) &= 
                \frac{\sum_{\mathbf{h}} P\left(\mathbf{x}, \mathbf{y},
                \mathbf{h}\right)}
                {\sum_{\mathbf{y}^{*}} \sum_{\mathbf{h}} P\left(\mathbf{x},
                \mathbf{y}^{*}, \mathbf{h}\right)} \\
            &= \frac{\sum_{\mathbf{h}} \exp\left(-E\left(\mathbf{x},
                \mathbf{y}, \mathbf{h}\right)\right)}{\sum_{\mathbf{y}^{*}}
                \sum_{\mathbf{h}} \exp\left(-E\left(\mathbf{x}, \mathbf{y}^{*},
                \mathbf{h}\right)\right)} 
            \label{eq:p_y_given_x}
        \end{aligned}
        \end{equation}
        where $\mathbf{y}$ is the one-hot encoding of a class label $y$,
        and $-\log \sum_{\mathbf{h}} \exp(-E(\mathbf{x}, \mathbf{y},
        \mathbf{h}))$ is the Free Energy $E_{free}$ of the RBM. 
        We consider the term containing the summation over $\mathbf{h}$ in
        (\ref{eq:p_y_given_x}):
        \begin{equation}
            \begin{aligned}
            \exp\left(E_{free}(\mathbf{x}, \mathbf{y})\right) 
            &= -\sum_{\mathbf{h}} \exp\left(-E\left(\mathbf{x}, \mathbf{y},
                \mathbf{h}\right)\right) \\ 
            &= -\sum_{\mathbf{h}} \exp\left(\sum_{i,j} x_i w_{ij} y_j + \sum_j
                u_{yj}h_j  + \sum_i a_i x_i + b_y + \sum_j c_j h_j\right) \\
            &= -\exp\left(\sum_i a_i x_i + b_y\right) 
                \sum_{\mathbf{h}} \exp\left(\sum_j h_j \sum_i x_i w_{ij} +
                u_{yj} + c_j\right) 
            \label{eq:free_energy}
            \end{aligned}
        \end{equation}
        Now consider only the second term of the product in
        (\ref{eq:free_energy}).
        We simplify it by re-writing $\sum_i x_i w_{ij} + u_{yj} + c_j$ as
        $\alpha_j$.
        Thus, we have
        \begin{equation}
            \begin{aligned}
            \sum_{\mathbf{h}} \exp\left(\sum_j h_j \sum_i x_i w_{ij} + u_{yj} +
                c_j\right) 
            &= \sum_{\mathbf{h}} \exp\left(\sum_j h_j \alpha_j\right) \\
            &= \sum_{\mathbf{h}} \prod_j \exp\left(h_j \alpha_j\right) \\
            &= \prod_j \sum_k \exp\left(s_k \alpha_j\right)
            \label{eq:transform}
        \end{aligned}
        \end{equation}
        where $s_k$ is each of the $k$ states that can be assumed by each
        hidden unit $j$ of the model. The last step of (\ref{eq:transform})
        results from re-arranging the terms after expanding the summation and
        product over $\mathbf{h}$ and $j$ in the previous step respectively. 
        The summation $\sum_\mathbf{h}$ over all the possible hidden layer
        vectors $\mathbf{h}$ can be replaced by the summation $\sum_k$ over the
        states of the units in the layer.
        The number and values of these states depend on the nature of the
        distribution in question (for instance $\{0, 1\}$ in the original
        DRBM).
        The result in (\ref{eq:transform}) can be applied to
        (\ref{eq:free_energy}) and, in turn, to (\ref{eq:p_y_given_x}) to get
        the following general expression of the conditional probability
        $P(y|\mathbf{x})$:
        \begin{equation}
        \begin{aligned}
            P(y|\mathbf{x}) 
            &= \frac{\exp\left(b_y\right) \prod_j \sum_k \exp\left(s_k
                \alpha_j\right)}{\sum_{y^{*}} \exp\left(b_{y^{*}}\right) 
                \prod_j \sum_k \exp\left(s_k \alpha^*_j\right)} \\
            &= \frac{\exp\left(b_y\right) \prod_j \sum_k \exp\left(s_k 
                \sum_i x_i w_{ij} + u_{yj} + c_j\right)}{\sum_{y^{*}}
                \exp\left(b_{y^*}\right) \prod_j \sum_k \exp\left(s_k  
                \sum_i x_i w_{ij} + u_{y^*j} + c_j\right)}
            \label{eq:drbm_generalisation}
        \end{aligned}
        \end{equation}
        The result in (\ref{eq:drbm_generalisation}) generalises the
        conditional probability of the DRBM first introduced in
        \cite{Larochelle2008}.
        The term inside the summation over $k$ can be viewed as a product
        between $\alpha_j$ corresponding to each hidden unit $j$ and each
        possible state $s_k$ of this hidden unit.
        Knowing this makes it possible to extend the original DRBM to be
        governed by other types of distributions in the hidden layer.

        \subsection{Extensions to other Hidden Layer Distributions}
        \label{subsec:drbm_extensions}
        We first use the result in (\ref{eq:drbm_generalisation}) to derive the
        expression for the conditional probability $P(y|\mathbf{x})$ in the 
        original DRBM \cite{Larochelle2008}.
        This will be followed by its extension, first to the $\{-1,
        +1\}$-Bernoulli distribution (referred to here as the \textit{Bipolar
        DRBM})and then the Binomial distribution (the \textit{Binomial DRBM}).   
        Section \ref{sec:experiments} presents a comparison between
        the performance of the DRBM with these different activations.

            \subsubsection{DRBM:}
            \label{subsubsec:logsig_act}
            The $\{0, 1\}$-Bernoulli DRBM corresponds to the model originally
            introduced in \cite{Larochelle2008}.
            In this case, each hidden unit $h_j$ can either be a $0$ or a
            $1$, i.e.  $s_k = \{0, 1\}$.
            This reduces $P(y|\mathbf{x})$ in
            (\ref{eq:drbm_generalisation}) to
            \begin{equation}
        \begin{aligned}
                P_{\textrm{ber}}\left(y|\mathbf{x}\right) 
                &= \frac{\exp\left(b_y\right) \prod_j \sum_{s_k \in \{0, 1\}}
                    \exp\left(s_k \alpha_j\right)} {\sum_{y^{*}}
                    \exp\left(b_{y^{*}}\right) \prod_j \sum_{s_k \in \{0, 1\}}
                    \exp\left(s_k \alpha^*_j\right)} \\
                &= \frac{\exp\left(b_y\right) \prod_j \left(1 +
                    \exp\left(\alpha_j\right)\right)} {\sum_{y^{*}}
                    \exp\left(b_{y^{*}}\right) \prod_j \left(1 +
                    \exp\left(\alpha^*_j\right)\right)} 
            \end{aligned}
        \end{equation}
            which is identical to the result obtained in \cite{Larochelle2008}.

            \subsubsection{Bipolar DRBM:}
            \label{subsubsec:tanh_act}
            A straightforward adaptation to the DRBM involves replacing its
            hidden layer states by $\{-1, +1\}$ as previously done in
            \cite{Freund1992} in the case of the RBM.
            This is straightforward because in both cases the hidden states of
            the models are governed by the Bernoulli distribution, however, in
            the latter case each hidden unit $h_j$ can either be a $-1$ or a
            $+1$, i.e. $s_k = \{-1, +1\}$.
            Applying this property to (\ref{eq:drbm_generalisation}) results in
            the following expression for $P(y|\mathbf{x})$:
            \begin{equation}
        \begin{aligned}
                P_{\textrm{bip}}\left(y|\mathbf{x}\right) 
                &= \frac{\exp\left(b_y\right) \prod_j \sum_{s_k \in \{-1, +1\}}
                    \exp\left(s_k \alpha_j\right)} {\sum_{y^{*}}
                    \exp\left(b_{y^{*}}\right) \prod_j\sum_{s_k \in \{-1, +1\}} 
                    \exp\left(s_k \alpha^*_j\right)} \\ 
                &= \frac{\exp\left(b_y\right) \prod_j
                    \left(\exp\left(-\alpha_j\right) +
                    \exp\left(\alpha_j\right)\right)} 
                    {\sum_{y^{*}} \exp\left(b_{y^{*}}\right) \prod_j
                    \left(\exp\left(-\alpha^*_j\right) +
                    \exp\left(\alpha^*_j\right)\right)}\ . 
            \end{aligned}
        \end{equation}

            \subsubsection{Binomial DRBM:}
            \label{subsubsec:bin_act}
            It was demonstrated in \cite{Teh2001} how groups of $N$ (where $N$
            is a positive integer greater than $1$) stochastic units of the
            standard RBM can be combined in order to approximate
            discrete-valued functions in its visible layer and hidden layers to
            increase its representational power.
            This is done by replicating each unit of one layer $N$ times and
            keeping the weights of all connections to each of these units from
            a given unit in the other layer identical.
            The key advantage for adopting this approach was that the learning
            algorithm remained unchanged.
            The number of these ``replicas'' of the same unit whose values are
            simultaneously $1$ determines the effective integer value (in the
            range $[0, N]$) of the composite unit, thus allowing it to assume 
            multiple values.
            The resulting model was referred to there as the Rate-Coded RBM
            (RBMrate).
            
            The intuition behind this idea can be extended to the DRBM by
            allowing the states $s_k$ of each hidden unit to assume integer
            values in the range $[0, N]$.
            The summation in (\ref{eq:drbm_generalisation}) would then be $S_N
            = \sum_{s_k=0}^N \exp\left(s_k \alpha_j\right)$, which simplifies
            as below 
            \begin{equation}
        \begin{aligned}
                S_N &= \sum_{s_k=0}^N \exp\left(s_k \alpha_j\right) \\
                &= 1 + \exp\left(\alpha_j\right)
                    \sum_{s_k=0}^{\left(N-1\right)} \exp\left(s_k
                    \alpha_j\right) \\ 
                &= \frac{1 - \exp\left(\left(N+1\right) \alpha_j\right)}{1 -
                    \exp\left(\alpha_j\right)} 
            \end{aligned}
        \end{equation}
            
            in (\ref{eq:drbm_generalisation}) to give
            \begin{equation}
        \begin{aligned}
                P_{\textrm{bin}}\left(y|\mathbf{x}\right) 
                &= \frac{\exp\left(b_y\right) \prod_j \sum_{s_k=0}^N
                    \exp\left(s_k \alpha_j\right)} {\sum_{y^{*}}
                    \exp\left(b_{y^{*}}\right) \prod_j \sum_{s_k=0}^N
                    \exp\left(s_k \alpha^*_j\right)} \\ 
                &= \frac{\exp\left(b_y\right) \prod_j
                    \frac{1-\exp\left(\left(N+1\right)
                    \alpha_j\right)}{1-\exp\left(\alpha_j\right)}} 
                    {\sum_{y^{*}} \exp\left(b_{y^{*}}\right) \prod_j
                    \frac{1-\exp\left(\left(N+1\right)
                    \alpha^*_j\right)}{1-\exp\left(\alpha^*_j\right)}}\ . 
            \end{aligned}
        \end{equation}

    \section{Experiments}
    \label{sec:experiments}
    We evaluated the Bipolar and the Binomial DRBMs on three benchmark machine
    learning datasets.
    These are two handwritten digit racognition datasets --- USPS
    \cite{Hastie2005} and MNIST \cite{Lecun1998}, and one document
    classification dataset --- 20 Newsgroups \cite{Lang1995}.
    Before going over the results of experiments carried out on each of these,
    we describe the evaluation methodology employed which was common to all
    three and the evaluation metric.

        \subsubsection{Methodology}
        \label{subsubsec:methodology_mnist}
        A grid search was performed to determine the best set of model
        hyperparameters. 
        The procedure involved first evaluating each of the trained models on a
        validation set and then selecting the best of these to be evaluated on
        the test set.
        When a dataset did not contain a pre-defined validation set, it was
        created using a subset of the training set.
        The initial learning rate $\eta_{init}$ for stochastic gradient descent
        was varied as $\{0.0001, 0.001, 0.01\}$.
        Early-stopping was used for regularisation.
        For this, the classification average loss of the model on the
        validation set was determined after every epoch.
        If the loss happened to be higher than the previous best one for ten
        consecutive epochs, the parameters were reverted back to their values
        in the previous best model, and training was resumed with a reduced
        learning rate.
        And if this happened five times, training was terminated. 
        The learning rate reduction was according to a schedule where it is
        progressively scaled by the factors $\frac{1}{2}$, $\frac{1}{3}$,
        $\frac{1}{4}$, and so on at each reduction step.
        The number of hidden units $n_{hid}$ was varied as $\{50, 100, 500,
        1000\}$.
        The maximum number of training epochs was set to 2000, but it was found
        that training always ended well before this limit.
        The negative log-likelihood error criterion was used to optimise the
        model parameters.
        The DRBM generated a probability distribution over the different
        classes.
        The class-label corresponding to the greatest probability value was
        chosen as the predicted class.
        As all three datasets contain only a single data split, i.e. only one
        set of training, validation and test sets, the results reported here
        are each an average over those obtained with $10$ model parameter
        initialisations using different randomisation seeds.
    
        An additional hyperparameter to be examined in the case of the Binomial
        DRBM is the number of bins $n_{bins}$.
        It corresponds to the number of states that can be assumed by each
        hidden unit of the model.
        The value of $n_{bins}$ was varied as $\{2, 4, 8\}$ in our experiments.

        \subsubsection{Evaluation Measure}
        \label{subsubsec:eval_mnist}
        In all of the prediction tasks, each model is expected to predict the
        one correct label corresponding to the image of a digit, or the
        category of a document.
        All the models in this task are evaluated using the average
        loss $E(\mathbf{y}, \mathbf{y}^*)$, given by: 
        \begin{equation}
            E(\mathbf{y}, \mathbf{y}^*) = \frac{1}{N} \sum_{i=1}^N
                \mathcal{I}\left(y_i \neq y^*_i\right)
        \end{equation}
        where $\mathbf{y}$ and $\mathbf{y}^*$ are the predicted and the
        true labels respectively, $N$ is the total number of test
        examples, and $\mathcal{I}$ is the $0-1$ loss function.

        \subsection{MNIST Handwritten Digit Recognition}
        \label{subsec:experiments_mnist}
        The MNIST dataset \cite{Lecun1998} consists of optical characters of
        handwritten digits.
        Each digit is a $28 \times 28$ pixel gray-scale image (or a vector
        $\mathbf{x} \in [0,1]^{784}$).
        Each pixel of the image corresponds to a floating-point value lying in
        the range $[0,1]$ after normalisation from an integer value in the
        range $[0,255]$.
        The dataset is divided into a single split of pre-determined training,
        validation and test folds containing $50,000$ images, $10,000$ images
        and $10,000$ images respectively.

        Table \ref{tab:drbm_mnist_results} lists the classification
        performance on this dataset of the three DRBM variants derived above
        using the result in (\ref{eq:drbm_generalisation}).
        The first row of the table corresponds to the DRBM introduced in
        \cite{Larochelle2008}.
        We did not perform a grid search in the case of this one model and
        only used the reported hyperparameter setting in that paper to
        reproduce their result\footnote{We obtained a marginally lower
        average loss of $1.78\%$ in our evaluation of this model than the
        $1.81\%$ reported in \cite{Larochelle2008}.}.
        It was stated there that a difference of $0.2\%$ in the average loss  
        is considered statistically significant on this dataset.
        \begin{table}[h]
            \centering
            \begin{tabular}{ll}
                \hline
                \textbf{Model} & \textbf{Average Loss(\%)} \\
                \hline
                DRBM ($n_{hid}=500$, $\eta_{init}=0.05$) & $\mathbf{1.78} (\pm 0.0012)$ \\
                Bipolar DRBM ($n_{hid}=500$, $\eta_{init}=0.01$) & $1.84 (\pm 
                    0.0007)$ \\ 
                Binomial DRBM ($n_{hid}=500$, $\eta_{init}=0.01$) & $1.86 (\pm
                    0.0016)$ \\
                \hline
            \end{tabular}
            \caption{A comparison between the three different variants of the
                DRBM on the USPS dataset.
                The Binomial DRBM in this table is the one with $n_{bins} = 2$.}
            \label{tab:drbm_mnist_results}
        \end{table}
        Going by this threshold of difference, it can be said that the
        performance of all three models is equivalent on this dataset although
        the average accuracy of the DRBM is the highest, followed by that of
        the Bipolar and the Binomial DRBMs.
        All three variants perform best with $500$ hidden units.
%        Going by the result in \cite{Teh2001}, it was expected that increasing
%        the value of $n_{bins}$ would positively influence the classification
%        performance.
%        This was, however, not the case.  
        It was observed that the number of bins $n_{bins}$ didn't play as
        significant a role as first expected.
        There seemed to be a slight deterioration in accuracy with an increase
        in the number of bins, but the difference cannot be considered
        significant given the threshold for this dataset.
        These results are listed in Table \ref{tab:bin_drbm_nbin_mnist}.
        \begin{table}[h]
            \centering
            \begin{tabular}{cccc}
                \hline
                $n_{bins}$ & $n_{hid}$ & $\eta_{init}$ & \textbf{Average Loss (\%)} \\ 
                \hline
                $2$ & $500$ & $0.01$ & $\mathbf{1.86}$ \\
                $4$ & $500$ & $0.01$ & $1.88$ \\
                $8$ & $500$ & $0.001$ & $1.90$ \\
                \hline
            \end{tabular}
            \caption{Classification performance of the Binomial DRBM with
                different values of $n_{bins}$ on the MNIST dataset.
                While the performance does show a tendency to worsen with
                the number of bins, the difference was found to be within the
                margin of significance for this dataset.}
            \label{tab:bin_drbm_nbin_mnist}
        \end{table}

        \subsection{USPS Handwritten Digit Recognition}
        \label{subsec:experiments_usps}
        The USPS dataset \cite{Hastie2005} contains optical characters of
        handwritten digits.
        Each digit is a $16 \times 16$ pixel gray-scale image (or a vector
        $\mathbf{x} \in [0,1]^{256}$).
        Each pixel of the image corresponds to a floating-point value lying in
        the range $[0,1]$ after normalisation from an integer value in the
        range $[0, 255]$.
        The dataset is divided into a single split of pre-determined training,
        validation and test folds containing $7,291$ images, $1,458$ images and
        $2,007$ images respectively.

        Table \ref{tab:drbm_usps_results} lists the classification
        performance on this dataset of the three DRBM variants derived above
        using the result in (\ref{eq:drbm_generalisation}).
        Here the Binomial DRBM (of $n_{bins}=8$) was found to have the best
        classification accuracy, followed by the Bipolar DRBM and then the DRBM.
        The number of hidden units used by each of these models varies
        inversely with respect to their average loss. 
        \begin{table}[h]
            \centering
            \begin{tabular}{ll}
                \hline
                \textbf{Model} & \textbf{Average Loss (\%)} \\
                \hline
                DRBM ($n=50$, $\eta_{init}=0.01$) & $6.90 (\pm 0.0047)$
                    \\ 
                Bipolar DRBM ($n=500$, $\eta_{init}=0.01$) & $6.49 (\pm 0.0026)$
                    \\  
                Binomial DRBM (8) ($n=1000$, $\eta_{init}=0.01$) &
                    $\mathbf{6.09} (\pm 0.0014)$ \\ 
                \hline
            \end{tabular}
            \caption{A comparison between the three different variants of the
                DRBM on the USPS dataset.
                The Binomial DRBM in this table is the one with $n_{bins} = 8$.}
            \label{tab:drbm_usps_results}
        \end{table}

        Table \ref{tab:bin_drbm_nbin_usps} shows the change in classification
        accuracy with a change in the number of bins. 
        In contrast to the observation in the case of MNIST, here an increase
        in $n_{bins}$ is accompanied by an improvement in accuracy.
        \begin{table}[h]
            \centering
            \begin{tabular}{cccc}
                \hline
                $n_{bins}$ & $\eta_{init}$ & $n_{hid}$ & \textbf{Average Loss
                    (\%)} \\  
                \hline
                $2$ & $0.01$ & $50$ & $6.90 (\pm 0.0047)$ \\
                $4$ & $0.01$ & $1000$ & $6.48 (\pm 0.0018)$ \\
                $8$ & $0.01$ & $1000$ & $\mathbf{6.09} (\pm 0.0014)$ \\
                \hline
            \end{tabular}
            \caption{Classification average losses of the Binomial DRBM with
                different values of $n_{bins}$.}
            \label{tab:bin_drbm_nbin_usps}
        \end{table}

        \subsection{20 Newsgroups Document Classification}
        \label{subsec:experiments_20news}
        The 20 Newsgroups dataset \cite{Lang1995} is a collection of
        approximately $20,000$ newsgroup documents, partitioned evenly across
        $20$ different categories. 
        A version of the dataset where the training and the test sets contain
        documents collected at different times is used here. 
        The aim is to predict the correct category of a document published
        after a certain date given a model trained on those published before
        the date.
        We used the $5,000$ most frequent words for the binary input features
        to the models.
        This preprocessing follows the example of \cite{Larochelle2008}, as it
        was the second data used to evaluate the DRBM there.
        We made an effort to adhere as closely as possible to the evaluation
        methodology there to obtain results comparable to theirs despite the
        unavailability of the exact validation set.
        Hence a validation set of the same number of samples was
        created\footnote{Our evaluation resulted in a model with a
        classification accuracy of $28.52\%$ in comparison with the $27.6\%$
        reported in \cite{Larochelle2008}.}.

        Table \ref{tab:drbm_20news_results} lists the classification
        performance on this dataset of the three DRBM variants derived above
        using the result in (\ref{eq:drbm_generalisation}).
        Here the Bipolar DRBM outperformed the remaining two variants, followed
        by the Binomial DRBM and the DRBM. 
        \begin{table}[h]
            \centering
            \begin{tabular}{ll}
                \hline
                \textbf{Model} & \textbf{Average Loss (\%)} \\
                \hline
                DRBM ($n=50$, $\eta_{init}=0.01$) & $28.52 (\pm 0.0049)$ 
                    \\ 
                Bipolar DRBM ($n=50$, $\eta_{init}=0.001$) & $\mathbf{27.75} (\pm 0.0019)$
                    \\  
                Binomial DRBM ($n=100$, $\eta_{init}=0.001$) & $28.17 (\pm 0.0028)$
                    \\ 
%                    ReLU-DRBM ($n=???$, $\eta_{init}=?.??$) & $?.??$ \\
                \hline
            \end{tabular}
            \caption{A comparison between the three different variants of the
                DRBM on the 20 Newsgroups dataset.
                The Binomial DRBM in this table is the one with $n_{bins} = 2$.}
            \label{tab:drbm_20news_results}
        \end{table}

        Table \ref{tab:bin_drbm_nbin_20news} shows the change in classification
        accuracy with a change in the number of bins. 
%        The best performance of models with a greater number of bins is better
%        than those with a smaller number of bins.
        \begin{table}[h]
            \centering
            \begin{tabular}{cccc}
                \hline
                $n_{bins}$ & $\eta_{init}$ & $n_{hidden}$ & \textbf{Average Loss
                    (\%)} \\  
                \hline
                $2$ & $0.001$ & $100$ & $\mathbf{28.17} (\pm 0.0028)$ \\
                $4$ & $0.001$ & $50$ & $28.24 (\pm 0.0032)$ \\
                $8$ & $0.0001$ & $50$ & $28.76 (\pm 0.0040)$ \\
                \hline
            \end{tabular}
            \caption{Classification performance of the Binomial DRBM with
                different values of $n_{bins}$.}
            \label{tab:bin_drbm_nbin_20news}
        \end{table}

    \section{Conclusions and Future Work}
    \label{sec:conclusions}
    This paper introduced a novel theoretical result that makes it possible to
    generalise the hidden layer activations of the Discriminative RBM (DRBM).
    This result was first used to reproduce the derivation of the cost function
    of the DRBM, and additionally to also derive those of two new variants of
    it, namely the Bipolar DRBM and the Binomial DRBM.  
    The three models thus derived were evaluated on three benchmark machine
    learning datasets --- MNIST, USPS and 20 Newsgroups.
    It was found that each of the three variants of the DRBM outperformed the
    rest on one of the three datasets, thus confirming that generalisations of 
    the DRBM may be useful in practice.

    It was found in the experiments in Section \ref{sec:experiments}, that
    the DRBM achieved the best classification accuracy on the MNIST dataset,
    the Bipolar DRBM on the 20 Newsgroups dataset and the Binomial DRBM on the
    USPS dataset.
    While this does indicate the practical utility of the two new variants of
    the DRBM introduced here, the question of whether each of these is better
    suited for any particular types of dataset than the rest is to be
    investigated further.%We leave this as future work. 

    Given the application of the result in (\ref{eq:drbm_generalisation}) to
    obtain the Binomial DRBM, it is straightforward to extend it to what we
    refer to here as the \textit{Rectified Linear DRBM}.
    This idea is inspired by \cite{Nair2010}, where the Rate-coded RBM
    \cite{Teh2001} (analogous to the Binomial DRBM here) is extended to derive
    an RBM with Rectified Linear units by increasing the number of replicas of
    a single binary unit to infinity.
    Adopting the same intuition here in the case of the DRBM, this
    would mean that we allow the states $s_k$ to assume integer values
    in the range $[0, \infty)$ and thus extend the summation $S_N$ in
    the case of the Binomial DRBM to an infinite sum $S_\infty$ resulting in
    the following derivation:
    \begin{equation}
        \begin{aligned}
        S_\infty 
        &= \sum_{s_k=0}^\infty \exp\left(s_k \alpha_j\right) \\
        &= 1 + \exp\left(\alpha_j\right) \sum_{s_k=0}^\infty
            \exp\left(s_k \alpha_j\right) \\ 
        &= \frac{1}{1 - \exp\left(\alpha_j\right)}
    \end{aligned}
        \end{equation}
    with the equation for the Rectified Linear DRBM posterior probability in 
    (\ref{eq:drbm_generalisation}) becoming
    \begin{equation}
        \begin{aligned}
        P_{\textrm{relu}}\left(y|\mathbf{x}\right) 
        &= \frac{\exp\left(b_y\right) \prod_j \sum_{s_k=0}^\infty
            \exp\left(s_k \alpha_j\right)} {\sum_{y^{*}}
            \exp\left(b_{y^{*}}\right) \prod_j \sum_{s_k=0}^\infty
            \exp\left(s_k \alpha^*_j\right)} \\ 
        &= \frac{\exp\left(b_y\right) \prod_j
            \frac{1}{1-\exp\left(\alpha_j\right)}} {\sum_{y^{*}}
            \exp\left(b_{y^{*}}\right) \prod_j
            \frac{1}{1-\exp\left(\alpha^*_j\right)}} \ .
    \end{aligned}
        \end{equation}
    Experiments with this variant of the DRBM are due, and will be carried out 
    in the future.

    % Bibliography
    \bibliographystyle{splncs03} % use this to have URLs listed in References
    \bibliography{references} % Path to your References.bib file

\begin{thebibliography}{10}
\providecommand{\url}[1]{\texttt{#1}}
\providecommand{\urlprefix}{URL }

\bibitem{Freund1992}
Freund, Y., Haussler, D.: {Unsupervised Learning of Distributions on Binary
  Vectors using Two Layer Networks}. In: Advances in Neural Information
  Processing Systems. pp. 912--919 (1992)

\bibitem{Glorot2010}
Glorot, X., Bengio, Y.: {Understanding the Difficulty of Training Deep
  Feedforward Neural Networks}. In: International Conference on Artificial
  Intelligence and Statistics. pp. 249--256 (2010)

\bibitem{Glorot2011}
Glorot, X., Bordes, A., Bengio, Y.: {Deep Sparse Rectifier Neural Networks}.
  In: International Conference on Artificial Intelligence and Statistics. pp.
  315--323 (2011)

\bibitem{Gupta1992}
Gupta, B.D., Schnitger, G.: {The Power of Approximation: A Comparison of
  Activation Functions}. In: Advances in Neural Information Processing Systems.
  pp. 615--622 (1992)

\bibitem{Hastie2005}
Hastie, T., Tibshirani, R., Friedman, J., Franklin, J.: {The Elements of
  Statistical Learning: Data Mining, Inference and Prediction}, chap.~1

\bibitem{Hinton2009}
Hinton, G., Salakhutdinov, R.: {Replicated Softmax: An Undirected Topic Model}.
  In: Advances in Neural Information Processing Systems. pp. 1607--1614 (2009)

\bibitem{Karlik2011}
Karlik, B., Olgac, A.V.: {Performance Analysis of Various Activation Functions
  in Generalized MLP Architectures of Neural Networks}. International Journal
  of Artificial Intelligence and Expert Systems  1(4),  111--122 (2011)

\bibitem{Lang1995}
Lang, K.: {Newsweeder: Learning to Filter Netnews}. In: Proceedings of the 12th
  international conference on machine learning. pp. 331--339 (1995)

\bibitem{Larochelle2008}
Larochelle, H., Bengio, Y.: {Classification using discriminative restricted
  Boltzmann machines}. In: International Conference on Machine Learning. pp.
  536--543. ACM Press (2008)

\bibitem{Lecun1998}
LeCun, Y., Bottou, L., Bengio, Y., Haffner, P.: {Gradient-based Learning
  Applied to Document Recognition}. Proceedings of the IEEE  86(11),
  2278--2324 (1998)

\bibitem{Lecun2006}
LeCun, Y., Chopra, S., Hadsell, R., Ranzato, M., Huang, F.: {A Tutorial on
  Energy-Based Learning}. Predicting Structured Data  (2006)

\bibitem{Lee2009}
Lee, H., Grosse, R., Ranganath, R., Ng, A.: {Convolutional Deep Belief Networks
  for Scalable Unsupervised Learning of Hierarchical Representations}. In:
  International Conference on Machine Learning. pp. 609--616. ACM (2009)

\bibitem{Mohamed2012}
Mohamed, A.R., Dahl, G., Hinton, G.: {Acoustic Modeling using Deep Belief
  Networks}. IEEE Transactions on Audio, Speech, and Language Processing
  20(1),  14--22 (2012)

\bibitem{Nair2010}
Nair, V., Hinton, G.: {Rectified Linear Units Improve Restricted Boltzmann
  Machines}. In: Proceedings of the 27th International Conference on Machine
  Learning (ICML-10). pp. 807--814 (2010)

\bibitem{Salakhutdinov2007}
Salakhutdinov, R., Mnih, A., Hinton, G.: {Restricted Boltzmann Machines for
  Collaborative Filtering}. In: Proceedings of the 24th international
  conference on Machine learning. pp. 791--798. ACM (2007)

\bibitem{Smolensky1986}
Smolensky, P.: Parallel distributed processing: Explorations in the
  microstructure of cognition, vol. 1. chap. {Information Processing in
  Dynamical Systems: Foundations of Harmony Theory}, pp. 194--281. MIT Press
  (1986)

\bibitem{Sutskever2007}
Sutskever, I., Hinton, G.: {Learning Multilevel Distributed Representations for
  High-Dimensional Sequences}. In: International Conference on Artificial
  Intelligence and Statistics. pp. 548--555 (2007)

\bibitem{Teh2001}
Teh, Y.W., Hinton, G.: {Rate-Coded Restricted Boltzmann Machines for Face
  Recognition}. Advances in Neural Information Processing Systems pp. 908--914
  (2001)

\bibitem{Tieleman2008}
Tieleman, T.: {Training Restricted Boltzmann Machines using Approximations to
  the Likelihood Gradient}. In: International Conference on Machine Learning.
  pp. 1064--1071. ACM (2008)

\bibitem{Welling2004}
Welling, M., Rosen-Zvi, M., Hinton, G.: {Exponential Family Harmoniums with an
  Application to Information Retrieval}. In: Advances in Neural Information
  Processing Systems. pp. 1481--1488 (2004)

\end{thebibliography}
\end{document}